\documentclass[conference]{IEEEtran}
\IEEEoverridecommandlockouts
\usepackage{amsmath,amssymb,amsfonts}
\usepackage{graphicx}
\usepackage{cite}
\usepackage{booktabs}
\usepackage{stfloats}
\usepackage{verbatim,subfig}
\title{Rethinking Skip Connections: Additive U-Net for Robust and Interpretable Denoising}
\author{\IEEEauthorblockN{Vikram R. Lakkavalli}
\IEEEauthorblockA{IIITB-Bangalore\\
Email: vikram.ramesh@iiitb.ac.in, vikram.ckm@gmail.com}
}
\begin{document}
\maketitle
\begin{abstract}
Skip connections are central to U-Net architectures for image denoising, but standard concatenation doubles channel dimensionality and obscures information flow, allowing uncontrolled noise transfer. We propose the \emph{Additive U-Net}, which replaces concatenative skips with gated additive connections. Each skip pathway is scaled by a learnable non-negative scalar, offering explicit and interpretable control over encoder contributions while avoiding channel inflation.

Evaluations on the Kodak-17 denoising benchmark show that Additive U-Net achieves competitive PSNR/SSIM at noise levels $\sigma=15,25,50$, with robustness across kernel schedules and depths. Notably, effective denoising is achieved even without explicit down/up-sampling or forced hierarchies, as the model naturally learns a progression from high-frequency to band-pass to low-frequency features.

These results position additive skips as a lightweight and interpretable alternative to concatenation, enabling both efficient design and a clearer understanding of multi-scale information transfer in reconstruction networks.
\end{abstract}

\section{Introduction}
Image denoising remains a central benchmark in low-level vision, testing an architecture’s ability to recover structure from corrupted inputs. Encoder–decoder networks such as U-Net \cite{ronneberger2015unet} dominate this space because skip connections allow fine details to bypass the bottleneck. Yet the standard \emph{concatenative} skip design introduces drawbacks: it doubles channel dimensionality, obscures which scale contributes to the output, and can transmit structured noise directly to the decoder---a phenomenon we term \emph{noise leakage}.

This raises a broader question: \emph{how does information flow across scales in denoising, and can we make this process interpretable?} Existing remedies such as residual shortcuts \cite{he2016resnet} or attention gating \cite{woo2018cbam} either add complexity or do not provide scalar, per-scale attribution.

We propose the \textbf{Additive U-Net}, which replaces concatenation with \textbf{gated additive} skips. Each skip pathway is scaled by a learnable non-negative coefficient, turning cross-scale fusion into an explicit, interpretable operation. This design has three key advantages:  
(i) \textbf{Simplicity}—no channel inflation or up/down-sampling;  
(ii) \textbf{Robustness}—skip weights constrain noise propagation; and  
(iii) \textbf{Interpretability}—learned scalars directly reveal the importance of each scale.

On the Kodak-17 benchmark with Gaussian noise ($\sigma=15,25,50$), Additive U-Net achieves PSNR/SSIM competitive with established baselines such as DnCNN, while remaining lightweight. Analysis of learned skip weights and filter responses shows how information is routed through scales, providing rare transparency in a denoising backbone. Beyond raw performance, our results highlight that \emph{controlling information flow may matter as much as depth}, with implications for interpretable design in medical imaging, scientific data processing, and multi-task learning.

\section{Related Work}

\subsection{Image denoising architectures}
Residual CNNs such as DnCNN \cite{zhang2017dncnn} established strong baselines for Gaussian denoising, later extended with dilation, non-local modules, and blind-noise training (e.g., FFDNet \cite{zhang2018ffdnet}, NLRN \cite{liu2018nonlocal}). Encoder–decoder designs such as U-Net \cite{ronneberger2015unet} remain popular for their skip-enabled detail recovery, though emphasis has been on maximizing PSNR/SSIM rather than analyzing feature flow.

\subsection{Skip connection designs}
Concatenative skips, standard in U-Net, improve fidelity but inflate channel dimensionality and obscure the role of individual scales. Alternatives include residual shortcuts (ResNet \cite{he2016resnet}), attention modules (CBAM \cite{woo2018cbam}), or conditional modulation layers (FiLM \cite{perez2018film}). These approaches provide flexibility but not explicit, scalar control of cross-scale contributions. Systematic evaluation of additive skips for denoising, with an emphasis on interpretability, remains largely unexplored.

\subsection{Multi-task extensions}
Auxiliary cues such as edges or segmentation have been used to assist restoration \cite{caruana1997multitask,chen2017hed,xu2017deepedgeguided}. While not our focus, the flat additive design naturally preserves the input domain, making it straightforward to attach auxiliary heads without entangling reconstruction and classification.

\noindent In summary, most prior work has treated skip connections as a performance device. Our work instead rethinks them as an \emph{explicit, interpretable information pathway} and evaluates additive skips in the denoising context.

\section{Proposed Method: Additive U-Net}
\label{sec:method}

\subsection{Design Motivation}
Our goal is a denoiser that (i) preserves a single feature space across depth, avoiding channel inflation from concatenation; (ii) omits down/up-sampling, reducing resolution-coupled biases; and (iii) provides explicit, interpretable control over skip contributions via learnable nonnegative scalars. This makes the model lightweight and transparent.

\begin{figure}[ht]
\centering
\includegraphics[scale=0.5]{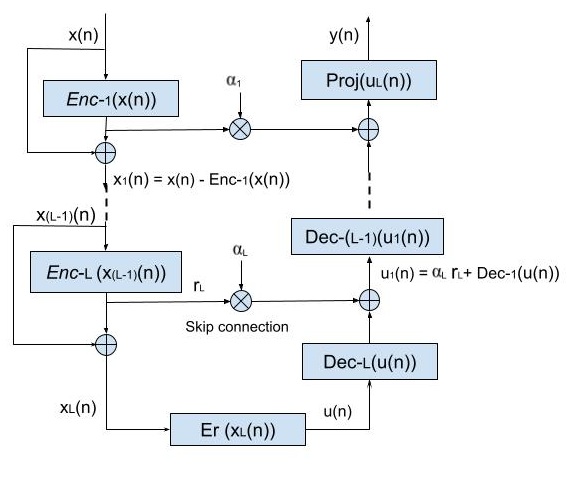}
\caption{Additive U-Net: each skip connection is scaled by a learnable coefficient $\alpha_j$, controlling information flow from encoder to decoder without channel concatenation. Corresponding encoder/decoder share same number of channels.}
\label{fig:unetArchitecture}
\end{figure}

\subsection{Architecture}
Let $x\in[0,1]^{1\times H\times W}$ be the noisy input. A $3{\times}3$ stem produces $x_0\!\in\!\mathbb{R}^{C\times H\times W}$.  

Each encoder block $\mathrm{Enc}_i$ predicts a residual $r_i$ and subtracts it from the current state:
\begin{equation}
r_i = \mathrm{Enc}_i(x_i), \qquad x_{i+1} = x_i - r_i.
\label{eq:enc}
\end{equation}
Residuals $\{r_1,\dots,r_L\}$ are cached as skip features.

The decoder fuses its running state $u_j$ with a gated skip from the matching encoder:
\begin{equation}
u_{j+1} = \mathrm{Dec}_j(u_j + \alpha_j r_{L-j}), \qquad \alpha_j \ge 0,
\label{eq:dec}
\end{equation}
where $\alpha_j=\mathrm{softplus}(\beta_j)$ ensures nonnegativity. Unlike concatenation, this additive fusion keeps channel width constant and makes each skip contribution scalar-explainable. A $1{\times}1$ head projects the final decoder state $u_L$ to the denoised output $y$.

\subsection{Loss and Training}
We train with the Charbonnier loss:
\begin{equation}
\mathcal{L}(y,\hat{x}) = \sqrt{(y-\hat{x})^2 + \epsilon^2}, \quad \epsilon=10^{-3},
\label{eq:charb}
\end{equation}
and report PSNR/SSIM on AWGN noise levels $\sigma \in \{15,25,50\}$.

\subsection{Variants}
We study kernel schedules $\{k_i\}$ such as \{3,3,3,3,3\}, \{1,3,5,7,9\}, and \{9,7,5,3,1\}, at depths $L\in\{3,5\}$. All models use the additive skip rule of Eq.~\eqref{eq:dec}, fixed channels $C$, and no down/up-sampling. All models use a fixed base channel width of 64 across layers, unless stated otherwise.

\section{Experimental Setup}
\label{sec:exp_setup}

We evaluate on the Kodak test set, a standard benchmark of $17$ high-quality $512{\times}768$ natural images ("Kodak-17"). Images are converted to grayscale and normalized to $[0,1]$ before corruption.

Additive white Gaussian noise (AWGN) with fixed standard deviation $\sigma \in \{15,25,50\}$ is added to form noisy inputs $x = y + n$, where $y$ is the clean reference and $n \sim \mathcal{N}(0, \sigma^2)$.

\noindent {Protocol:} We train on the combined DIV2K~\cite{Agustsson2017DIV2K} and Flickr2K~\cite{Timofte2017NTIRE} datasets. Following standard practice, we extract random grayscale crops of size $128{\times}128$ with $K$ independent noise realizations per clean patch to improve robustness. All models use the Adam optimizer with learning rate $2{\times}10^{-4}$, batch size $4$, and are trained for $200$ epochs. We employ the Charbonnier loss defined in Eq.~\eqref{eq:charb} with $\epsilon=10^{-3}$.

\subsection{Architectural Variants}
Unless otherwise specified, we use depth $L=5$ with base channel width $C=64$. We evaluate kernel schedules $\{k_i\}$ such as \{\texttt{3,3,3,3,3}\}, \{\texttt{1,3,5,7,9}\}, and \{\texttt{9,7,5,3,1}\}, as well as a shallow variant with depth $L=3$ (\{\texttt{3,3,3}\}). All models use the additive encoder/decoder rule of Sec.~\ref{sec:method}, with no downsampling or upsampling.

\subsection{Comparison Baselines}
To contextualize our Additive U-Net, we implement two representative baselines.

\paragraph{DnCNN.}
The DnCNN model~\cite{zhang2017dncnn} is a residual CNN denoiser without skip connections. It serves as a comparison to architectures that rely purely on depth and residual learning without explicit encoder-decoder symmetry.

\paragraph{Pseudo-additive U-Net.}
We also consider a "pseudo-additive" variant that mimics a conventional U-Net but replaces concatenation with direct addition of encoder and decoder features. Unlike our subtractive encoder (Eq.~\eqref{eq:enc}), this model passes encoder activations forward without residual bookkeeping. This baseline allows us to isolate the effect of true subtractive encoding and gated additive skips in our proposed model.

\subsection{Evaluation Metrics}
We report Peak Signal-to-Noise Ratio (PSNR, dB) and Structural Similarity (SSIM)~\cite{Wang2004SSIM} averaged over all 17 Kodak test images. We highlight both quantitative scores and qualitative examples to illustrate failure modes (e.g., line straightness, texture leakage).

\section{Results and Discussion}

\subsection{Quantitative Performance}
Table~\ref{tab:psnr} shows that Real Additive U-Net achieves PSNR/SSIM competitive with DnCNN and pseudo-additive baselines, despite its simpler design. Performance remains robust across kernel schedules and depths, confirming that additive skips avoid channel inflation while preserving effectiveness.

\subsection{Qualitative Comparisons}
\begin{figure*}[t]
  \centering
  \includegraphics[scale=0.4]{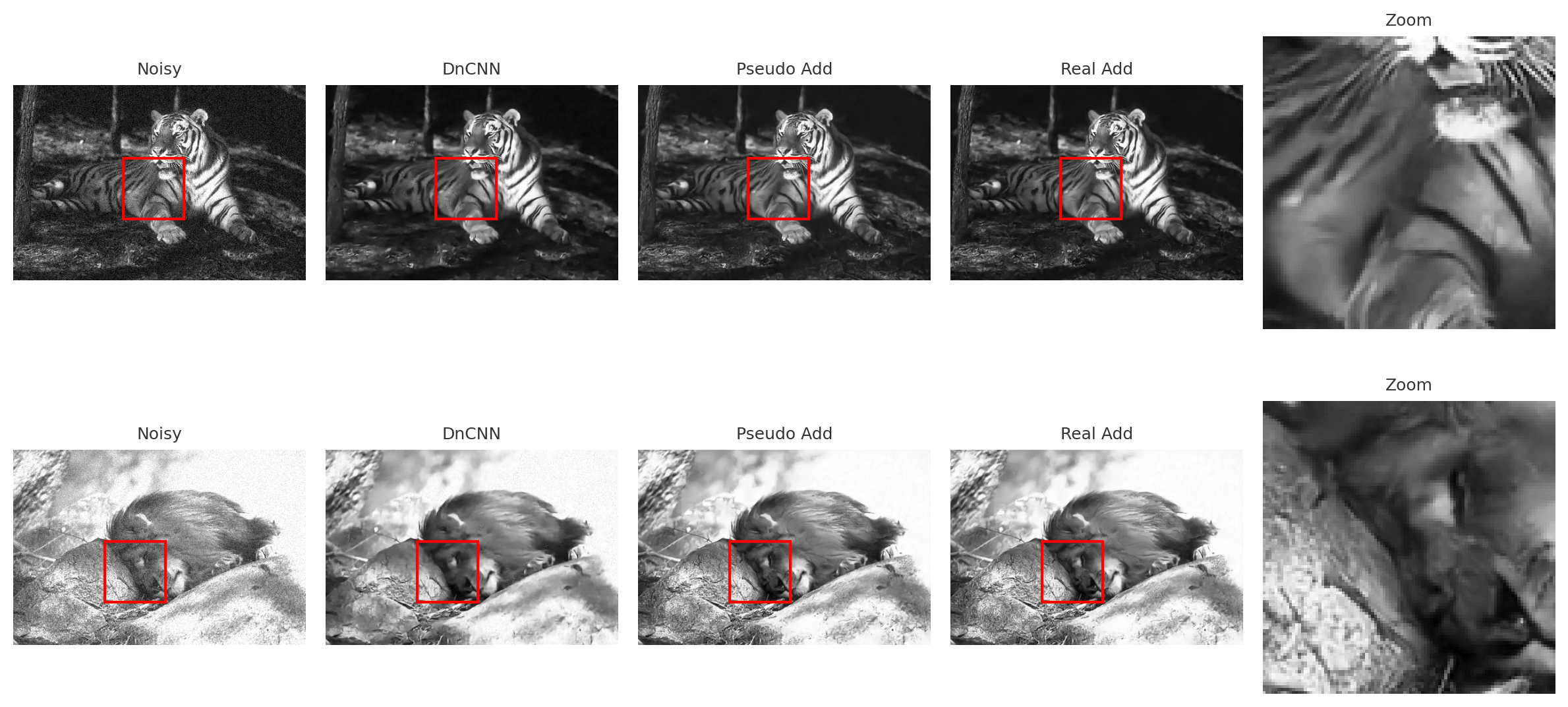}
  \caption{\textit{Visual comparison (Kodak-17, Test009 and Test005, $\sigma$=15). DnCNN oversmooths, Pseudo Add-U-Net leaves blotchy artifacts, while Real Add-U-Net preserves edges and natural textures.}}
  \label{fig:tiger}
\end{figure*}

Figure~\ref{fig:tiger} illustrates that DnCNN tends to oversmooth fine details, while pseudo-additive U-Net leaves blotchy artifacts. Real Additive U-Net produces sharper textures and cleaner edges, suppressing noise leakage without sacrificing natural structure. At high noise levels (Fig.~\ref{fig:noisy50}), it continues to preserve contours where DnCNN oversmooths.

\subsection{Interpretability}
\begin{figure}[t]
\centering
\includegraphics[scale=0.25]{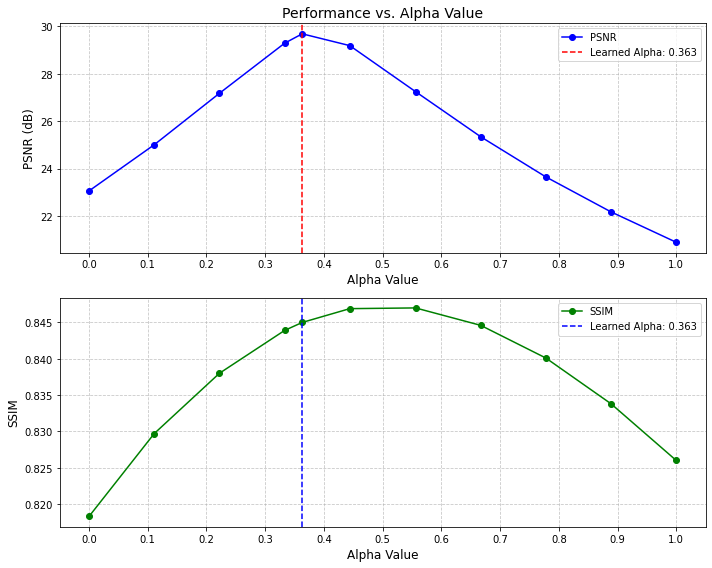}
\caption{\textit{Role of $\alpha$ in decoding. Varying $\alpha_3$ from 0–1 yields smooth changes in PSNR/SSIM, with peak near the learned value, confirming interpretable and optimized feature fusion.}}
\label{fig:alpha}
\end{figure}

\begin{figure}[t]
    \centering
    \subfloat[Radial frequency profiles across encoder layers]{
        \includegraphics[width=0.2\textwidth]{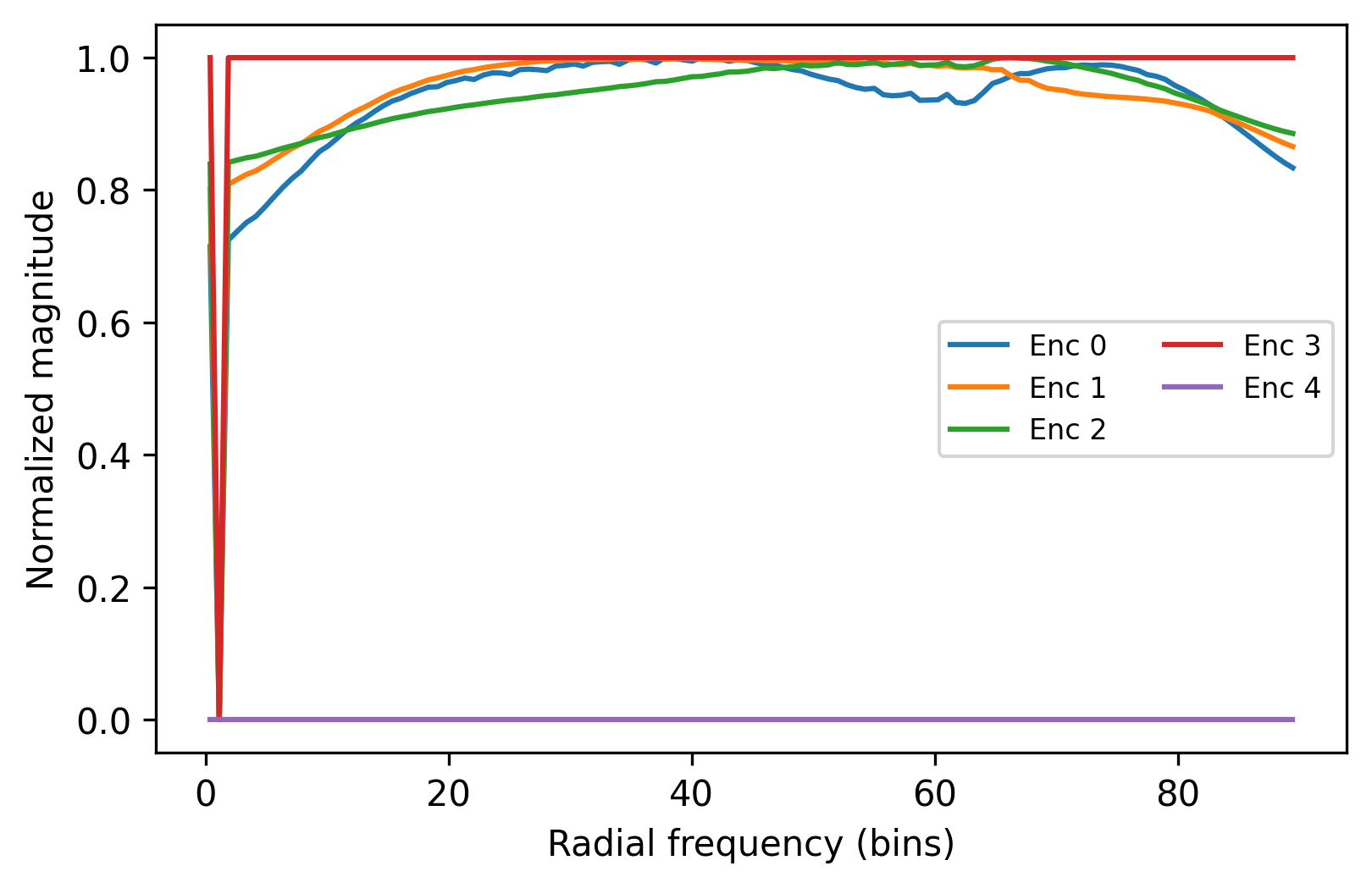}
    }
    \hfill
    \subfloat[Encoder 0: Top-$K$ filter FFT exemplars]{
        \includegraphics[width=0.2\textwidth]{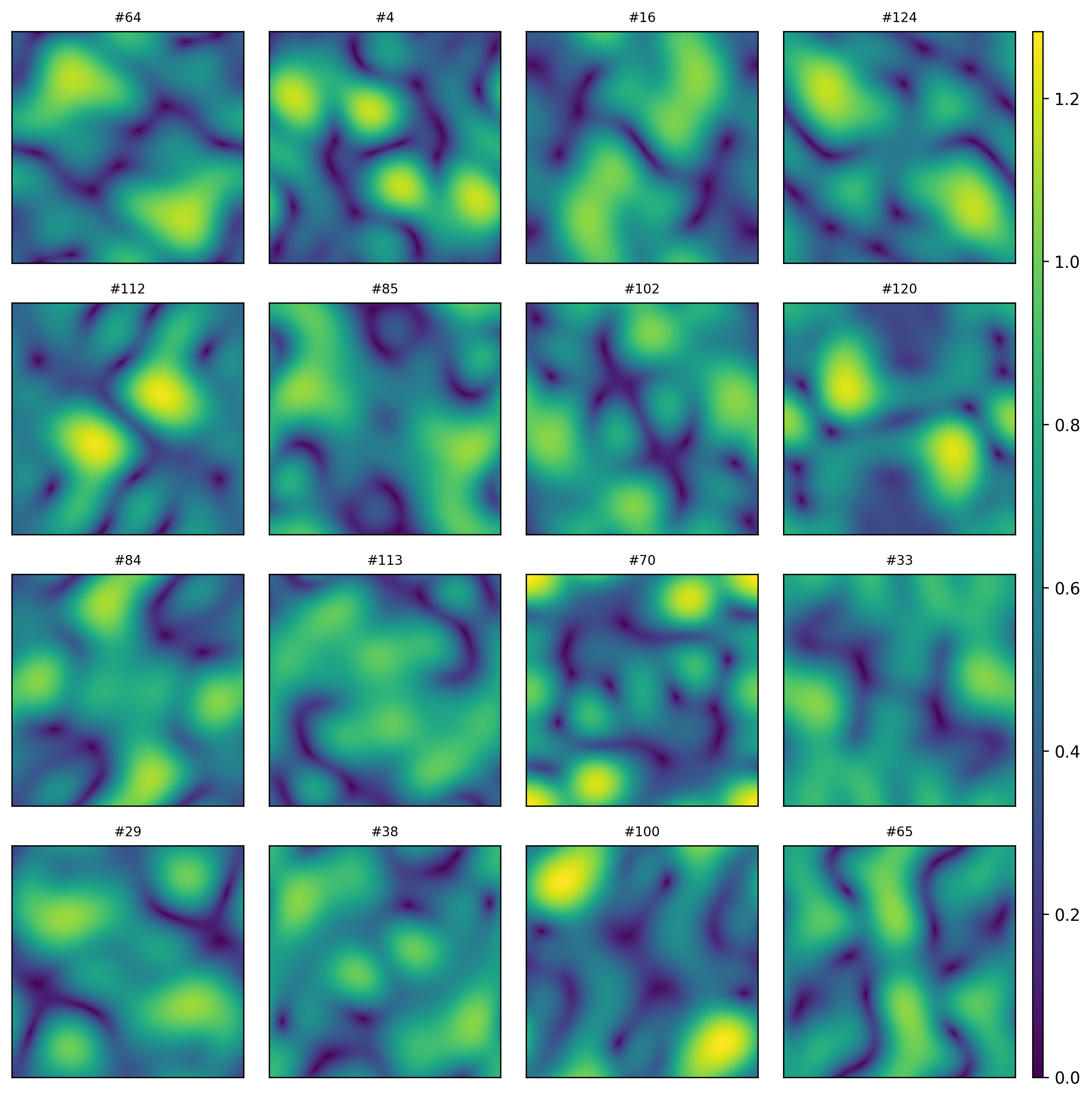}
    }
    \vspace{1em}
    \subfloat[Encoder 1: Top-$K$ filter FFT exemplars]{
        \includegraphics[width=0.2\textwidth]{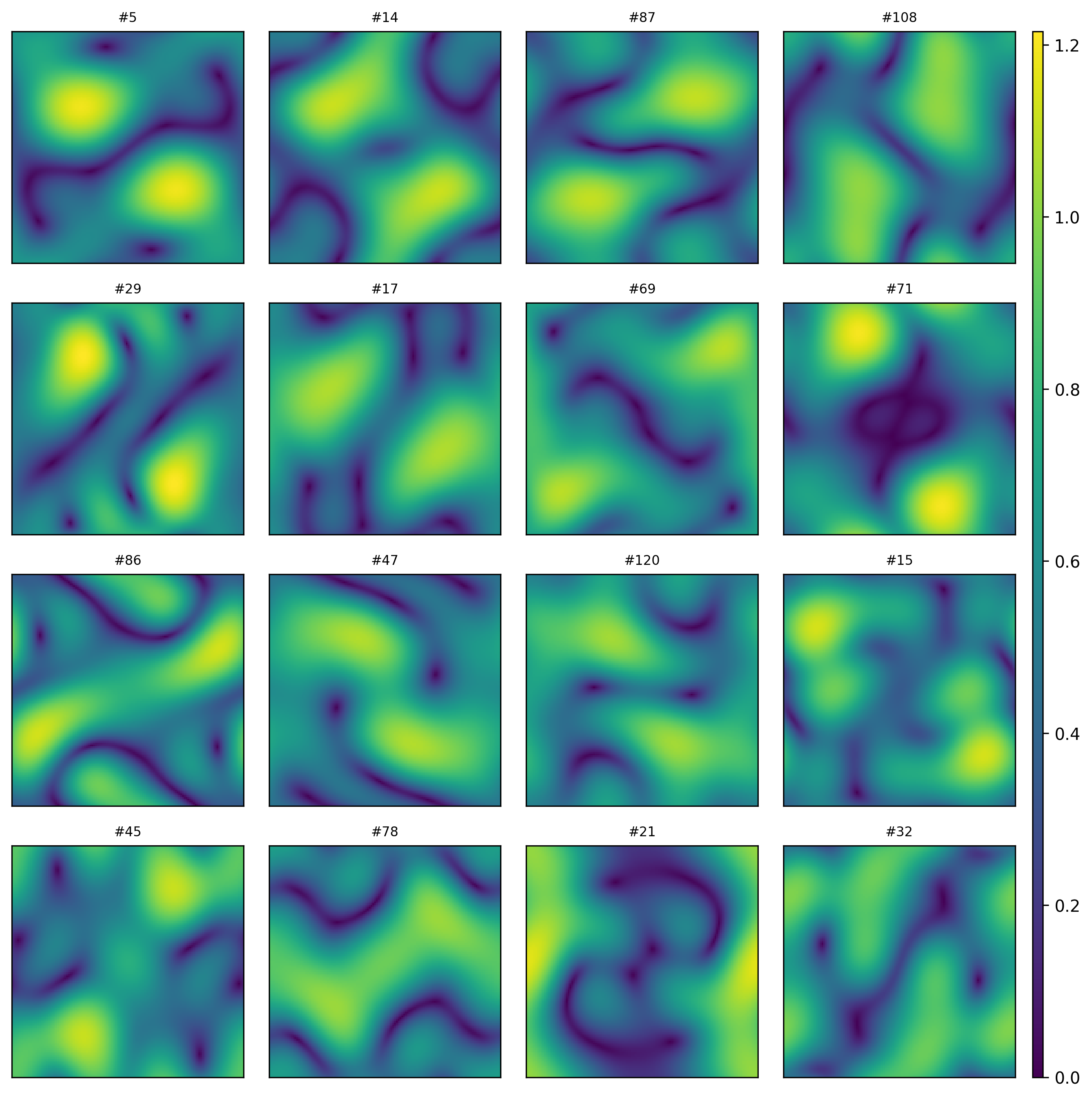}
    }
    \hfill
    \subfloat[Encoder 2: Top-$K$ filter FFT exemplars]{
        \includegraphics[width=0.2\textwidth]{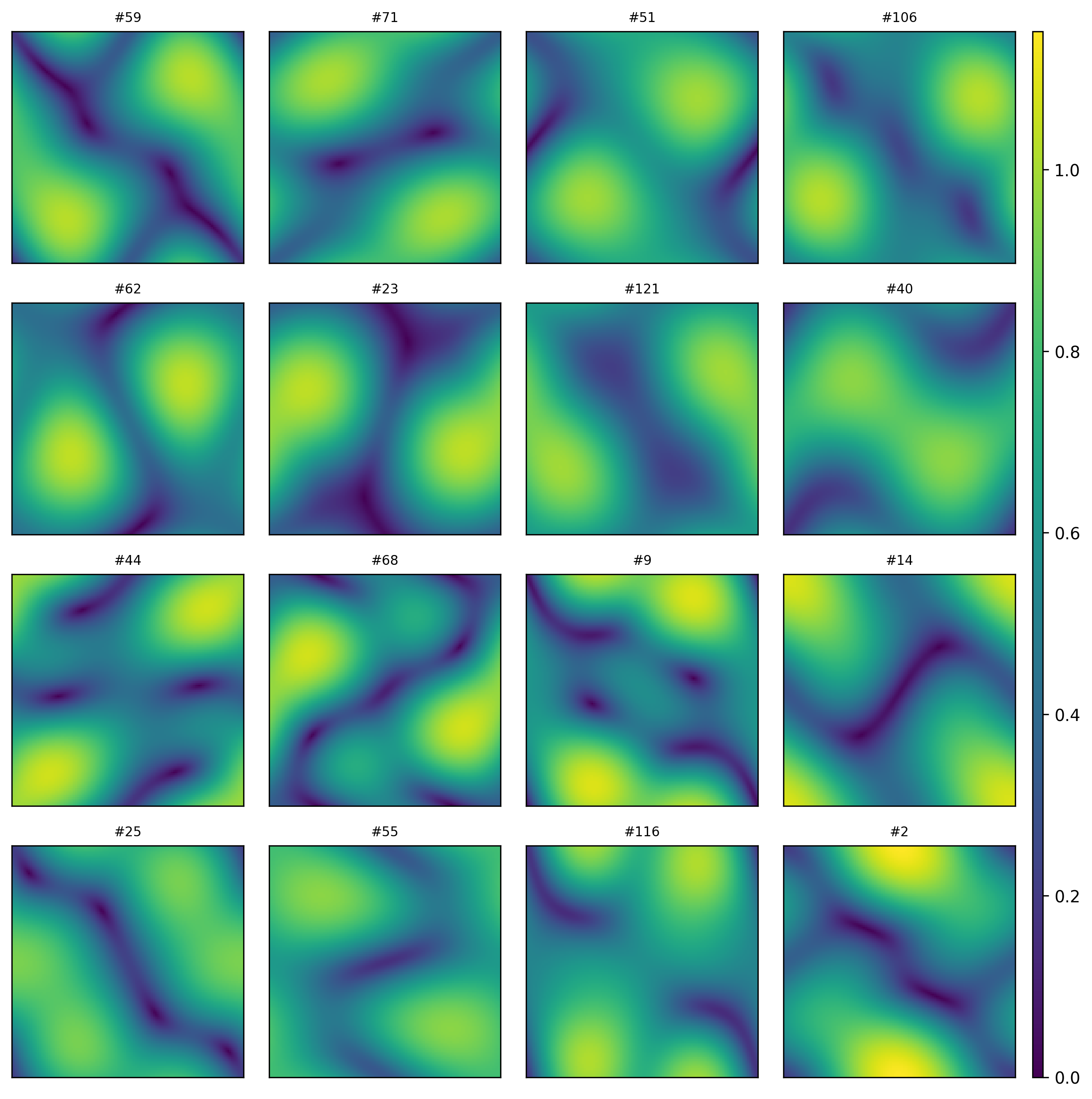}
    }
    \caption{\textit{Frequency analysis of filters. Encoder layers specialize from high- to low-frequency features, showing an emergent hierarchical progression without explicit multi-scale design.}}
    \label{fig:addunet_fft}
\end{figure}

\begin{figure}[t]
  \centering
  \includegraphics[width=\columnwidth]{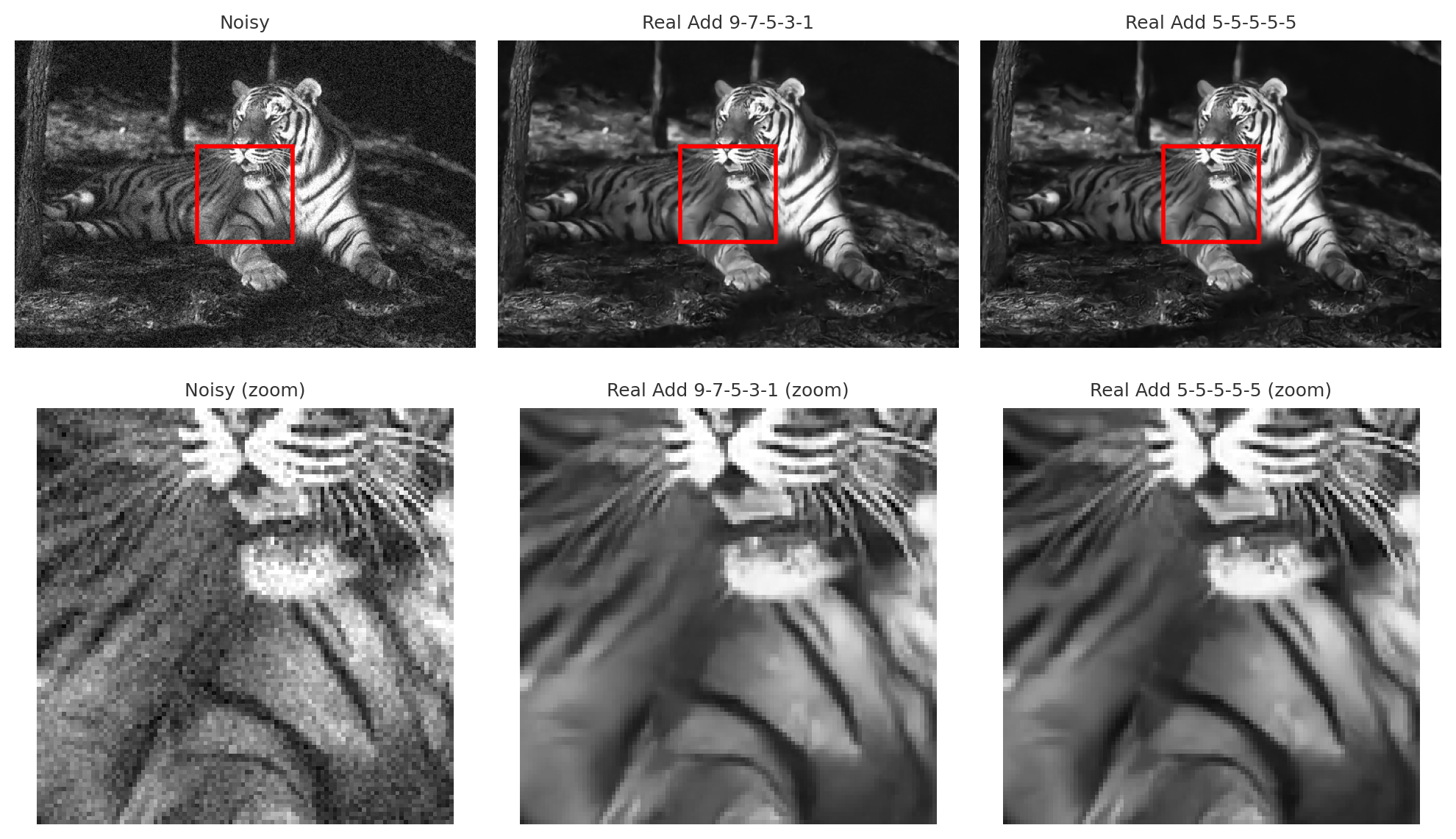}
  \caption{\textit{Kernel schedule comparison ($\sigma$=15). The {9–7–5–3–1} variant emphasizes fine detail but retains grain, while {5–5–5–5–5} produces smoother backgrounds and sharper global contrast.}}
  \label{fig:kernel-comparison}
\end{figure}

Skip gate sweeps (Fig.~\ref{fig:alpha}) demonstrate smooth, interpretable control of feature contributions, with peak performance near the learned values. Kernel schedule comparisons (Fig.~\ref{fig:kernel-comparison}) reveal that {9–7–5–3–1} emphasizes fine detail while {5–5–5–5–5} yields globally cleaner contrast. Frequency-domain analysis (Fig.~\ref{fig:addunet_fft}) further shows an emergent progression from high- to low-frequency features without explicit multi-scale design.

\begin{table}[ht]
\centering
\caption{Average PSNR (dB) / SSIM on Kodak-17 for Gaussian noise $\sigma \in \{15,25,50\}$. P-AddU represents the pseudo-additive U-Net baseline that uses direct feature addition without residual bookkeeping, while R-AddU represents our proposed Real Additive U-Net with subtractive encoding and gated skips. The results demonstrate that our approach achieves competitive performance across noise levels while maintaining architectural simplicity.}
\label{tab:psnr}
\begin{tabular}{lccc}
\toprule
Model (depth, kernels) & $\sigma{=}15$ & $\sigma{=}25$ & $\sigma{=}50$ \\
\midrule
P-AddU (5, 3-3-3-3-3) & 32.03 / 0.898 & 29.68 / 0.844 & 26.49 / 0.743 \\
R-AddU (5, 3-3-3-3-3)   & 31.53 / 0.888 & 29.21 / 0.833 & 25.34 / 0.709 \\
R-AddU (5, 5-5-5-5-5)   & 31.03 / 0.879 & 28.98 / 0.821 & 25.67 / 0.707 \\
R-AddU (5, 9-7-5-3-1)   & 31.73 / 0.896 & 28.88 / 0.828 & 25.23 / 0.719 \\
R-AddU (5, 1-3-5-7-9)   & 30.16 / 0.887 & 27.46 / 0.827 & 24.00 / 0.721 \\
R-AddU (3, 3-3-3)       & 31.63 / 0.891 & 29.34 / 0.836 & 26.31 / 0.732 \\
\bottomrule
\end{tabular}
\end{table}

We also present the role of $\alpha$ in deciding the output of the U-Net denoiser. Figure~\ref{fig:alpha} shows the variation of PSNR and SSIM when $\alpha_3$ is varied.


Figure~\ref{fig:addunet_fft} illustrates the frequency-domain analysis of our model.

\begin{figure}[t]
    \centering
    \subfloat[Noisy input, $\sigma{=}50$]{%
        \includegraphics[width=0.24\linewidth]{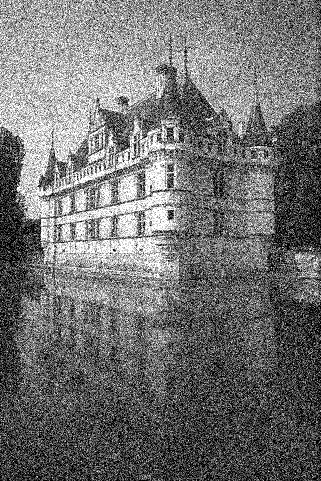}}
    \hfill
    \subfloat[Real Additive U-Net (97531)]{%
        \includegraphics[width=0.24\linewidth]{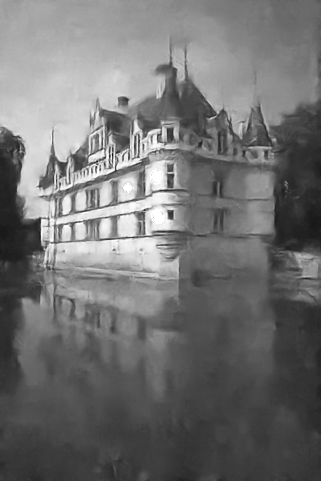}}
    \hfill
    \subfloat[Real Additive U-Net (33333)]{%
        \includegraphics[width=0.24\linewidth]{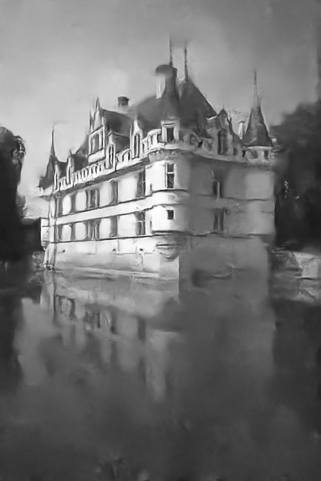}}
    \hfill
    \subfloat[DnCNN-17]{%
        \includegraphics[width=0.24\linewidth]{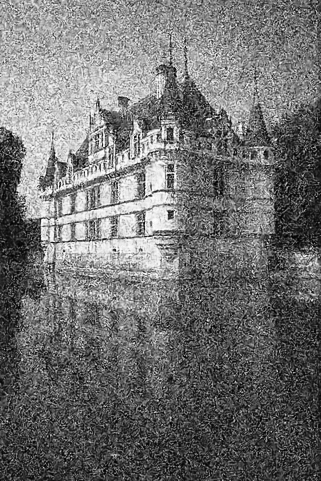}}
    \caption{%
        Visual comparison of denoising performance at $\sigma{=}50$. 
        The Real Additive U-Net preserves structural details such as edges and contours 
        while reducing structured noise leakage. Two kernel schedules are compared: 
        (b) uneven 97531 and (c) uniform 33333. Both yield competitive results, with 
        slight differences in smoothness and edge sharpness. 
    }
    \label{fig:noisy50}
\end{figure}

Figure~\ref{fig:noisy50} shows visual comparison at high noise level $\sigma=50$.

\section{Conclusion}
\label{sec:conclusion}
We presented the Additive U-Net, a lightweight encoder–decoder architecture that replaces concatenative skips with gated additive connections and subtractive encoding. Experiments on Kodak-17 demonstrate that this design achieves competitive PSNR/SSIM across noise levels while avoiding channel inflation and mitigating noise leakage.

Beyond raw performance, additive skips offer interpretable control of feature fusion, with learned gate values aligning with optimal contributions. Frequency analysis further shows that the model automatically organizes features from high- to low-frequency across depth, even without explicit multi-scale design.

Finally, the flat additive design naturally preserves the input domain, making it straightforward to attach auxiliary heads for multi-task settings (e.g., joint denoising and segmentation). Exploring this in future work could further highlight the versatility of additive skips.

These results highlight that controlling information flow can matter as much as depth, suggesting additive skips as a practical and transparent alternative for denoising. Future work may extend this principle to attention or spectral blocks, enabling interpretable design in medical imaging and scientific reconstruction tasks.

\bibliographystyle{IEEEbib}

\end{document}